\documentclass[conference]{IEEEtran}
\IEEEoverridecommandlockouts
\usepackage{cite}
\usepackage{amsmath,amssymb,amsfonts}
\usepackage{algorithmic}
\usepackage{graphicx}
\usepackage{hyperref}
\usepackage{textcomp}
\usepackage{xcolor}
\usepackage{comment}
\def\BibTeX{{\rm B\kern-.05em{\sc i\kern-.025em b}\kern-.08em
    T\kern-.1667em\lower.7ex\hbox{E}\kern-.125emX}}
\begin{document}

\title{HCT: Hybrid Convnet-Transformer for Parkinson's disease detection and severity prediction from gait\\

\thanks{}
}

\author{\IEEEauthorblockN{Safwen Naimi}
\IEEEauthorblockA{\textit{Data Science Laboratory} \\
\textit{University of Quebec (TÉLUQ)}\\
Montréal, Canada} \\
\and
\IEEEauthorblockN{Wassim Bouachir}
\IEEEauthorblockA{\textit{Data Science Laboratory} \\
\textit{University of Quebec (TÉLUQ)}\\
Montréal, Canada} \\
\and
\IEEEauthorblockN{Guillaume-Alexandre Bilodeau}
\IEEEauthorblockA{\textit{LITIV lab.} \\
\textit{Polytechnique Montréal}\\
Montréal, Canada}\\
}

\maketitle

\begin{abstract}
In this paper, we propose a novel deep learning method based on a new Hybrid ConvNet-Transformer architecture to detect and stage Parkinson's disease (PD) from gait data. We adopt a two-step approach by dividing the problem into two sub-problems. 
Our Hybrid ConvNet-Transformer model first distinguishes healthy versus parkinsonian patients. If the patient is parkinsonian, a multi-class Hybrid ConvNet-Transformer model determines the Hoehn and Yahr (H\&Y) score to assess the PD severity stage.
Our hybrid architecture exploits the strengths of both Convolutional Neural Networks (ConvNets) and Transformers to accurately detect PD and determine the severity stage. 
In particular, we take advantage of ConvNets to capture local patterns and correlations in the data, while we exploit Transformers for handling long-term dependencies in the input signal. We show that our hybrid method achieves superior performance when compared to other state-of-the-art methods, with a PD detection accuracy of 97\% and a severity staging accuracy of 87\%. Our source code is available at \href{https://github.com/SafwenNaimi/HCT-Hybrid-Convnet-Transformer-for-Parkinson-s-disease-detection-and-severity-prediction-from-gait}{https://github.com/SafwenNaimi}.
\end{abstract}


\section{Introduction}
Parkinson's disease (PD) is a progressive neurological condition that impairs movement control. It manifests itself by tremors, rigidity, and difficulties with coordination and balance. PD is caused by the death of dopamine-producing cells in the \textit{substantia nigra}, which is a part of the brain responsible for movement regulation \cite{Hughes1994ParkinsonsDA}. The specific cause of Parkinson's disease is unknown, although research shows that hereditary and environmental factors may play an important role. The diagnosis is mainly based on a combination of clinical symptoms and signs, and can be difficult since symptoms can be similar to those of other neurological conditions. The Hoehn and Yahr scale (H\&Y) \cite{Hoehn1998ParkinsonismOP} is often used to categorize Parkinson's disease into five stages based on the severity of the symptoms. The Unified Parkinson's Disease Rating Scale (UPDRS) \cite{Fahn1987UnifiedPD} is another widely used tool for evaluating the severity of symptoms. Both scales are used in clinical practice and research to monitor the progression of the disease and evaluate the outcomes of interventions.

Gait information can be used to detect distinctive abnormalities in the walking patterns of PD patients, such as decreased step length and increased step variability. The advantage of using gait data is the ease of practical implementation compared to other types, such as speech data \cite{Vilda2017ParkinsonDD, Amato2021AnAF}. Additionally, the calculation of quantitative gait parameters and the extraction of informative gait features are of paramount importance, as they can be used to calculate significant clinical spatiotemporal parameters, such as swing phase, stance phase, and stride time. These gait parameters have been associated with H\&Y stages, and some studies have shown that gait measures can be utilized to predict a patient's H\&Y stage. For PD detection from gait data, several machine learning methods have been proposed and achieved promising results \cite{Zhao2018AHS, Ertugrul2016DetectionOP, Wu2017MeasuringSF}. This includes deep neural networks, multi-layer perceptron, random forest, and support vector machines. Despite the widespread use of these methods, most of them are only limited to binary classification for determining if a patient has PD based on gait information. However, PD staging, which corresponds to the detection of  severity stage is another classification problem that is less explored by the research community.

In this work, we present a new approach for detecting Parkinson's disease and estimating the severity stage using gait data. Our method adopts a two-step strategy that divides the problem of diagnosing PD into a detection step and a staging step. In the first step, the framework detects the presence of PD. If PD is detected, the second step involves determining the stage of the disease. The main contribution of this work is a new hybrid deep learning architecture exploiting the capabilities of ConvNets and Transformers to diagnose PD in a two-step process. We take advantage of ConvNet and Transformer architectures in the two steps to extract both local and global features respectively from gait information captured from foot sensors. The ConvNet captures local patterns, while the Transformer captures long-term dependencies and temporal relationships in the 1D signal. Thus, our model can learn complex relationships in the data, which are indicative of PD. Our experiments demonstrate that the proposed approach is more accurate than existing methods for detecting and staging PD.

\section{Related Work}
Vertical Ground Reaction Force (VGRF) signal has been widely employed for the diagnosis and classification of PD since it has been demonstrated to be a critical and discriminative kinematic parameter in PD detection and staging \cite{Guo2022DetectionAA}. 
Perumal and Sankar \cite{Perumal2016GaitAT} utilized a linear discriminant analysis (LDA)-based pattern classification algorithm for early detection and monitoring of PD using gait and tremor features. They achieved an average accuracy of 86.9\% in identifying PD tremors by analyzing frequency domain characteristics. However, their approach is limited to detecting the presence of PD, without identifying the stage of the disease. Aşuroğlu et al. \cite{Aurolu2018ParkinsonsDM} proposed a locally weighted random forest regression model to handle the effects of interpatient variability in gait features. In their work, VGRF sensor data are used to model the relationships between gait patterns and PD symptoms. They provided a quantitative assessment of PD motor symptoms using 16 time-domain and 7 frequency-domain features. However, only the statistical analysis of VGRF was used for identifying PD symptoms, and the kinematic analysis and severity level of PD were not reported. El Maachi et al. \cite{Machi2019Deep1F} proposed a 1D-ConvNet to extract features from gait data and detect PD, which has been shown to significantly improve PD detection results. The work of Nguyen et al. \cite{Nguyen2022TransformersF1} implemented a combination of Temporal Transformer and Spatial Transformer models to detect Parkinson's disease without addressing the staging task, which also yielded promising results. Another study from Veeraragavan et al. \cite{Veeraragavan2020ParkinsonsDD} used a Feed Forward Network (FFN) to analyze gait data and detect the H\&Y stages of PD patients. 

These methods were proven to be useful in diagnosing PD based on gait data. Most of them focus on classifying the subject as healthy or parkinsonian, while a few address the staging task. However, relying only on FFNs to collect local information may not be sufficient since they are limited in capturing the complex patterns in physiological data. Similarly, relying only on 1D-ConvNets to capture sensor associations may not be the best choice, as they are better suited for capturing local spatial information rather than global patterns or correlations among different sensors. Transformers, used in \cite{Nguyen2022TransformersF1}, are good at capturing global relationships between data, but are not as good to process local information. This is due to their architecture that focuses primarily on long-range dependencies, which may not allow capturing fine-grained and local patterns present in physiological data. These observations motivated our approach to incorporate Convolutional Neural Network (ConvNet) and Transformer architectures in a single model for PD diagnosis. Our work tackles these limitations by proposing a new Hybrid Convnet-Transformer (HCT) architecture leveraging the strengths of both architectures. Our HCT architecture is used according to a two-step strategy, which entails detecting the presence and then accurately predicting the stage of the disease. 
\begin{figure}[t]
    \centering
	\includegraphics[width=0.35\textwidth]{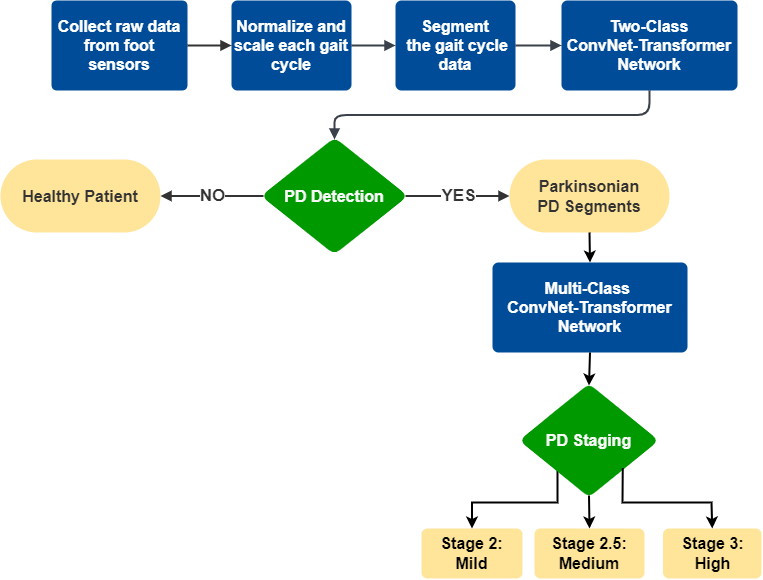}
	\caption{Flowchart of our overall algorithm for diagnosing PD}
	\label{FIG:3}
\end{figure}
\begin{figure*}[h]
\centering
\includegraphics[width=0.79\textwidth]{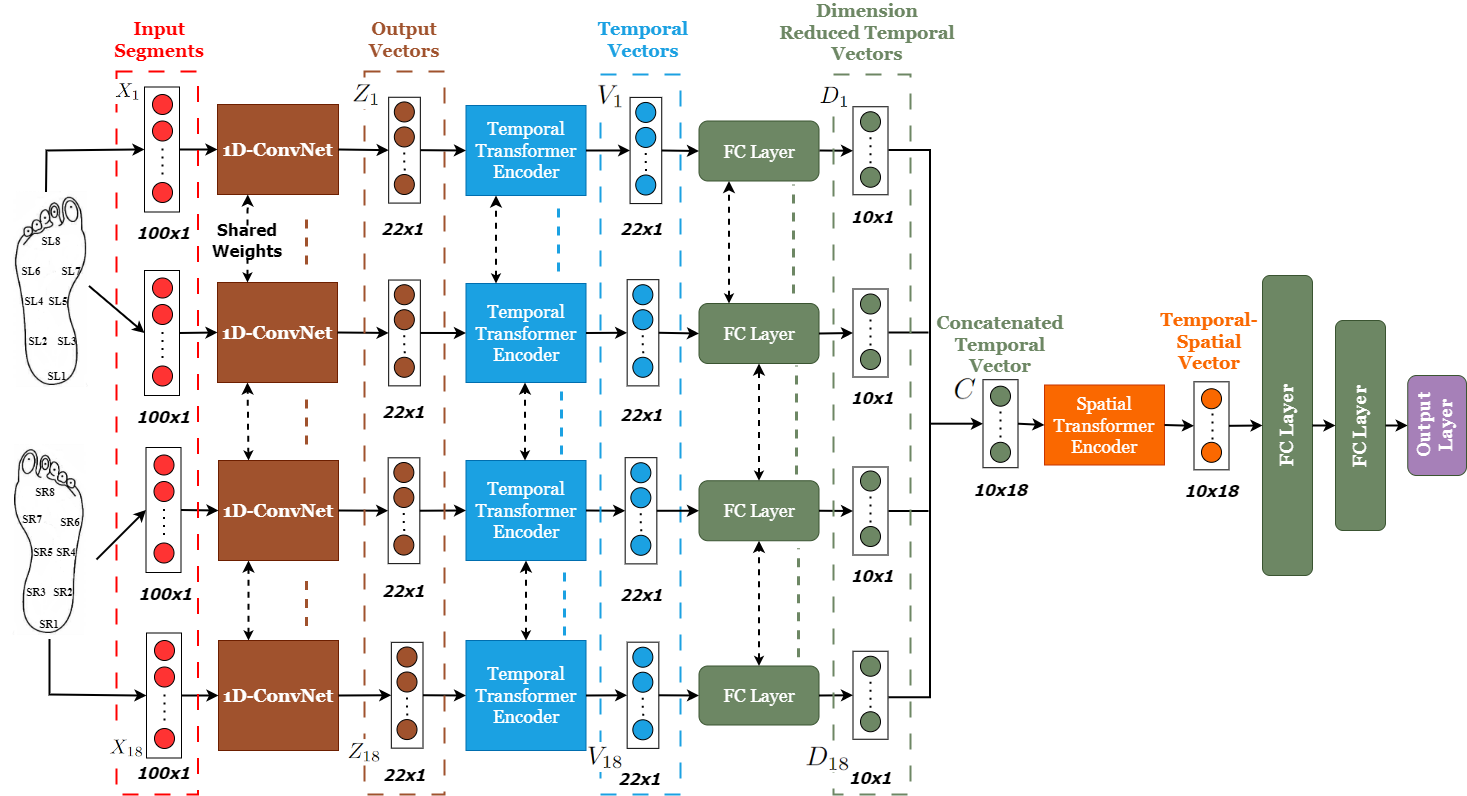}
\caption{The proposed Hybrid Convnet-Transformer (HCT) architecture.  For the PD detection step, we used this architecture with an output layer comprising a single neuron for binary classification (Healthy or PD). This model is named the Two-class ConvNet-Transformer. For the PD staging model, we used this architecture with three neurons on the output layer (stages 2, 2.5, and 3). This model is named the Multi-class ConvNet-Transformer.}
\label{fig:label}
\end{figure*}
\begin{figure}[b]
\centering
\includegraphics[width=0.4\textwidth]{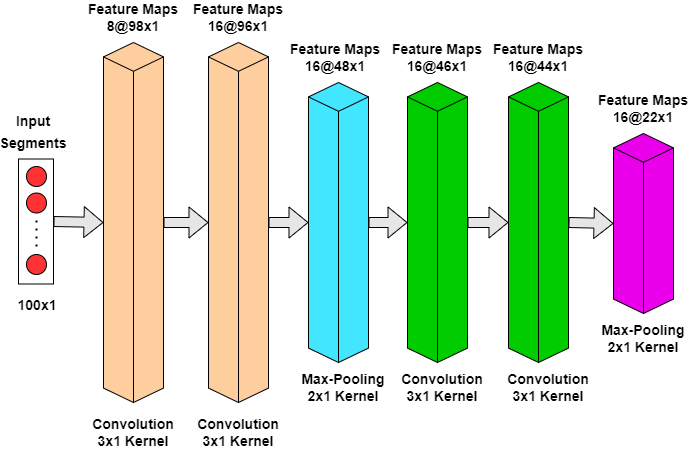}
\caption{Internal structure of the 1D-ConvNet network}
\label{fig:conv1d}
\end{figure}
\section{Method}
The flowchart of our overall algorithm for diagnosing PD is presented in Figure \ref{FIG:3}. We divide this problem into two steps. First, we detect the presence of disease from gait signals derived from foot sensors. The output of the Two-class ConvNet-Transformer Network, in that case, is a binary prediction of either a healthy patient or a parkinsonian patient. If the patient is parkinsonian, the same gait signal is passed to another Multi-Class ConvNet-Transformer Network to predict the exact stage of PD. The details of our method are presented in the following subsections.
\subsection{Data preprocessing}
We start by inputting 1-Dimensional VGRF signals from a patient's walk provided by foot sensors denoted as $s_i(t)$, where $t$ represents time and $i$ is a sensor index. These signals represent the vertical ground reaction force measured in Newtons as a function of time recorded by the foot sensors. During preprocessing, we replace missing H\&Y values of control patients with zeroes and normalize the signals to have a mean of zero and a standard deviation of one using
\begin{equation}
s_i(t) \leftarrow \frac{s_i(t) - \mu_s}{\sigma_s} ;
\end{equation}
where $\mu_s$ and $\sigma_s$ represent the mean and standard deviation of the signal $s_i(t)$ respectively. Finally, we segment the signals into distinct segments of $n$ elements with $n=100$ elements in our case. The segment is denoted as
 \begin{equation}
X_i^j = [x^j_i(1), x_i^j(2), ..., x_i^j(n)];    
\end{equation}
where $i$ is the sensor index ($i \in [1, 18]$), $j$ is the segment index, $n$ is the number of elements. The 100-time step is chosen so that enough data is maintained in each segment for characterizing the walk, while keeping the vectors short enough to enable the model to function without requiring too much memory.

\subsection{Hybrid ConvNet-Transformer architecture}

The detailed architecture of our Hybrid ConvNet-Transformer (HCT) model (Two-class or Multi-class) is provided in Figure \ref{fig:label}. The first part is made up of 18 parallel 1D-Convnets with shared parameters. Each ConvNet accepts the segments $X_i^j$ of size $100 \times 1$ elements produced from the 1D-VGRF signals as input and processes it through four convolutional layers. A max-pooling layer comes after every two convolutional layers as shown in Figure \ref{fig:conv1d}.
This 1D-Convnets parallelization allows the processing of each signal separately. In fact, each time series has its own deep features and is evaluated individually using the 1D-Convnet parallelization, since each sensor collects different data from a specific sensor position. The 1D-ConvNet is applied to each segment $X_i^j$ independently, and the output is represented by the vector $Y_i^j$ of size $22 \times 1$ elements given by
\begin{equation}
Y_i^j = [y^j_i(1), y_i^j(2), ..., y_i^j(k)];    
\end{equation}
where $i$ is the sensor index ($i \in [1, 18]$), $j$ is the segment index, $k$ is the number of elements. 

The temporal dependencies, which are the connections between two values of a vector spaced apart in time, can be captured using the temporal transformers. We incorporate a fixed positional encoding into the output of each 1D ConvNet, as the Transformer architecture does not consider the order of elements in a sequence. The choice of a fixed positional encoding is related to signal segmentation into a fixed 100-element vector size before inputting them into the 1D-ConvNet layers. The positional encodings are the element position in the 1D-ConvNet output vector.
The fixed positional encoding is given by
 \begin{equation}
P = [p(1), p(2), ..., p(k)] ;   
\end{equation}
 where $p(k)$ represents the position of the element $k$ in $Y_i^j$. The final output after adding the positional encoding is denoted by:
 \begin{equation}
     Z_i^j = Y_i^j + P ;
 \end{equation}
This output $Z_i^j$ from the $i^{th}$ sensor for the $j^{th}$ segment contains 22 elements. It is processed through a temporal transformer encoder block. This encoder block consists of a multi-head attention layer with four heads and a feed-forward network, which is similar to the architecture proposed in BERT for natural language processing \cite{Vaswani2017AttentionIA}. The result of this process is a temporal vector $V_i^j$ that also contains 22 elements.

After reducing the dimensions of the vectors from 22 elements to 10 elements using Fully Connected (FC) Layers, we obtain dimension-reduced temporal vectors of size $10 \times 1$ elements represented as
\begin{equation}
D_i^j = [d^j_i(1), d_i^j(2), ..., d_i^j(l)];    
\end{equation}
where $l$ is the number of elements.

These vectors are then concatenated to increase computational efficiency, resulting in a tensor $C$ of size $10 \times 18$ elements. This resulting tensor is then used as input to the encoder of the spatial transformer, with the addition of a fixed positional encoding. This positional encoding provides the spatial transformer encoder with information about the relative positions of the input elements in the concatenated vector. The sensor index is used as positional encodings. This spatial transformer is used to capture the spatial dependencies between each set of vectors coming from the 18-foot sensors. The spatial transformer encoder block used in our architecture is also made of one multi-head attention layer with four heads and a feed-forward network.

\subsection{Output layer}

The last component of our Hybrid ConvNet-Transformer model is a fully connected network operating on the concatenation of the features extracted by the spatial transformer encoder. For PD detection using our Two-class ConvNet-Transformer Network, the output layer is made of a single neuron with a Sigmoid activation function, which produces a probability of the input segment $X_i^j$ belonging to the class of PD. We used the binary cross-entropy loss defined as
\begin{equation}
L_{b} = -\frac{1}{M} \sum_{m=1}^{M} [a_m \log(\hat{a_m}) + (1 - a_m) \log(1 - \hat{a_m})];
\end{equation}
where $M$ is the number of samples, $a_m$ is the true label, and $\hat{a_m}$ is the predicted probability of the input belonging to the positive class (PD).

For PD staging using our Multi-class ConvNet-Transformer Network, the output layer is made of three neurons with a Softmax activation function to produce the probability of each of the three severity stages. In this case, the categorical cross-entropy loss \cite{LeCun1999ObjectRW} is used as
\begin{equation}
L_{c} = -\frac{1}{M} \sum_{m=1}^{M} \sum_{b=1}^{B} d_{m,b} \log(\hat{d}_{m,b});
\end{equation}
where $M$ is the number of samples, $B$ is the number of classes, $d_{m,b}$ is the one-hot encoding of the true label for the $m-th$ sample and the $b-th$ class, and $\hat{d}_{m,b}$ is the predicted probability of the $m-th$ sample belonging to the $b-th$ class.
The final walk classification in both cases is decided according to the majority classification of all the subject walk segments. 
\section{Experiments}
This section presents the experiments conducted to evaluate the performance of our proposed approach. Here, we describe the dataset used, the evaluation metrics, and the implementation details.  We finally provide and discuss the results.
\subsection{Dataset Description}
To evaluate the proposed Hybrid ConvNet-Transformer architecture, we used the PhysioNet gait dataset \cite{Physionet}.
The dataset was created by three groups of researchers, namely 
Yogev et al. \cite{Yogev2005DualTG}, Hausdorff et al. \cite{ Hausdorff2007RhythmicAS} and Toledo et al. \cite{FrenkelToledo2005TreadmillWA}. The dataset contains the gait patterns from 93 patients affected with PD and 73 healthy subjects. It has three gait patterns acquired through walking on level ground, walking with rhythmic auditory stimulation (RAS), and walking on a treadmill. Therefore, in total, 300 walks have been recorded from the aforementioned 166 individuals. 210 (70 \%) of the recorded walks are parkinsonian and 90 (30\%) control walks. There are a total number of 18-time series signals for each walk, including 16 (8x2) VGRF signals recorded from 8 sensors on each foot, and 2 additional total VGRFs under each foot.

\subsection{Evaluation Metrics}
We tested our algorithm using 10-fold cross-validation. At the subject level, we separated the Parkinson's and the control groups into 10 folds. As a result, we were able to maintain the same dataset balance for each fold (70\% Parkinson - 30\% Control).
We use the following notations to present the evaluation metrics: $TP$ as the number of true positives, $TN$ as the number of true negatives, $FP$ as the number of false positives, and $FN$ as the number of false negatives.
Our Two-class ConvNet-Transformer model dedicated to PD detection was evaluated using specificity (Sp), sensitivity (Se), and accuracy (Acc). These metrics are calculated with
\begin{equation}
\text { \bf Sensitivity: } S e=\frac{T P}{T P+F N}; \text { \bf Specificity: } S p=\frac{T N}{T N+F P} ;
\end{equation}
and
\begin{equation}
\text { \bf Accuracy (\%): } Acc =\frac{TP+TN}{TP+TN+FP+FN}\times 100\%;
\end{equation}

For our Multi-class ConvNet-Transformer model dedicated to PD staging, we report the precision, recall, and F1 score for each class using
\begin{equation}
\text { \bf Precision: } Pr=\frac{TP}{TP+FP}; \text { \bf Recall: } Re=\frac{TP}{TP+FN};
\end{equation}
and
\begin{equation}
\text { \bf F1-Score } =2\times \frac{ Pr \times Re}{ Pr+ Re};
\end{equation}

\subsection{Training details}
The proposed Hybrid ConvNet-Transformer models (Two-class and Multi-class) were trained, validated, and tested separately, using a batch size of 200 samples for each iteration. We trained for 25 epochs. Every fully connected layer, except for the output layer, uses the SeLU activation function (scaled exponential linear units) \cite{Pratama2020TrainableAF}. The proposed model is trained using the Nadam stochastic optimization method. We adjusted the parameters of our optimizers as follows: $\alpha_B$ = 0.0005, $\alpha_M$ = 0.001, $\beta_1$ = 0.9, and $\beta_2$ = 0.999, where $\alpha_B$ is the learning rate for the Two-class ConvNet-Transformer model, $\alpha_M$ is the learning rate for the Multi-class ConvNet-Transformer model, $\beta_1$ and $\beta_2$ are the exponential decay rates for the first and second-moment estimations, respectively.
To improve the performance of our model and reduce overfitting, we opted for a dropout rate of 0.3 and early stopping.
\begin{figure}[b]
     \centering
		\includegraphics[scale=0.47]{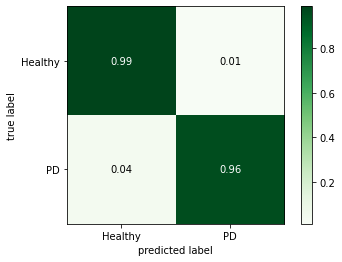}
	\caption{Confusion matrix of the Two-class HCT model}
	\label{FIG:101}
\end{figure}

\subsection{Results}
This section presents the results of our method for diagnosing Parkinson's disease from gait data in a two-step process. We evaluate the performance of our approach in detecting PD and determining the severity of the disease, and we provide an overall performance evaluation of the entire framework.
\begin{table}[ht]
   \centering
  \caption{Cross-validation results for binary classification of PD patients and healthy controls. The best results are in bold. Sp: Specificity, Se: Sensitivity, Acc: Accuracy, SD: Standard deviation
}
\begin{tabular}{|c|c|c|c|}
\hline
   \bf{\textit{Method}} & \bf{\textit{Se(\%) $\pm$ SD}} & \bf{\textit{Sp(\%) $\pm$ SD}} & \bf{\textit{Acc(\%) $\pm$ SD}} \\
\hline
        GLRA \cite{Wu2017MeasuringSF} &  n/a          &   n/a         &  82.8          \\
\hline
        SVM \cite{Wu2017MeasuringSF} &  n/a          &   n/a         &  84.5          \\
\hline
Random Forest \cite{Ertugrul2016DetectionOP} &        n/a &        n/a &       86.9 \\
\hline
       MLP \cite{Ertugrul2016DetectionOP} &       88.9 &       82.2 &       88.9 \\
\hline
       DNN \cite{Zhao2018AHS} & 96.2 $\pm$3.8 & 76.7 $\pm$ 8.2 & 90.3 $\pm$2.9 \\
\hline
1D-ConvNet \cite{Machi2019Deep1F} &  97.0 $\pm$ 4 & \textbf{88.5} $\pm$ 11.3 & 94.5 $\pm$5.2 \\
\hline
Transformer \cite{Nguyen2022TransformersF1} & 98.1 $\pm$ 3.2 & 86.8 $\pm$ 8.2 & 95.2 \bf{$\pm$ 2.3} \\
\hline
Two-class HCT (ours) & \bf{98.7 $\pm$ 2.3}  & 86.1 \bf{$\pm$ 5.6}  & \textbf{97.0} $\pm$ 2.7 \\

\hline

\end{tabular}  
 \label{tab:tab55}%
\end{table}%
\subsubsection{PD Detection results}
The performance of our Two-class ConvNet-Transformer model dedicated to PD Detection is compared with other methods in Table \ref{tab:tab55}. This table provides 10-fold cross-validation results for Parkinson's detection using various methods. It presents the three performance metrics for each method with the corresponding standard deviation (SD) over the 10 folds. Our method achieved the best results in terms of Accuracy and Sensitivity. However, there was a slight decrease in Specificity compared to 1D-ConvNet \cite{Machi2019Deep1F} and the Transformer method \cite{Nguyen2022TransformersF1}, but our method is more consistent with lower standard deviation for Sensitivity and Specificity. This demonstrates that exploiting ConvNets and Transformers in our model leads to stable results across multiple folds, outperforming 1D-ConvNets or Transformers when used alone in \cite{Machi2019Deep1F} and \cite{Nguyen2022TransformersF1}, respectively. The smaller standard deviation in Sensitivity and Specificity reflects that our model has gained stability across different folds during cross-validation. The confusion matrix of our Two-class ConvNet-Transformer model is shown in Figure \ref{FIG:101}.

\begin{figure}[b]
     \centering
		\includegraphics[scale=0.47]{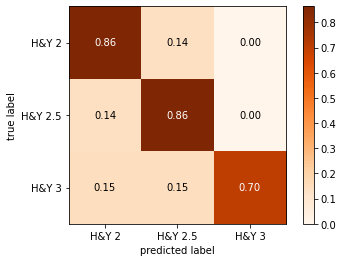}
	\caption{Confusion matrix of the Multi-class HCT model}
	\label{FIG:102}
\end{figure}

\subsubsection{PD Staging results}
\begin{table}[ht]
   \centering
  \caption{Comparison of Parkinson's Severity Assessment Algorithms. The best results are in bold}
\begin{tabular}{|c|c|c|c|c|}
\hline \bf{\textit{Method}}  & \bf{\textit{Accuracy}}  &
\bf{\textit{Precision}}  &
\bf{\textit{Recall}}  &
\bf{\textit{F1-Score}}\\
\hline FFN \cite{Veeraragavan2020ParkinsonsDD}  & $86.50 \%$ & $\mathbf{87.73\%}$ & $87.55 \%$ & $87.67 \%$ \\
\hline Multi-Class HCT (Ours) & $\mathbf{87.10\%}$ & $86.05 \%$ & $\mathbf{88.52\%}$ & $\mathbf{88.03\%}$ \\
\hline
\end{tabular}
  \label{tab:tab54}%
\end{table}%
\begin{table*}[ht]
   \centering
  \caption{Comparison of multiple methods for PD diagnosis. The best result is in bold}
\begin{tabular}{|c|c|c|c|c|c|}
\hline \bf{\textit{Method}} & \bf{\textit{Architecture}}  & \bf{\textit{Accuracy}} &
\bf{\textit{Precision}}  &
\bf{\textit{Recall}}  &
\bf{\textit{F1-Score}}\\
\hline Caramia et al. \cite{Caramia2018IMUBasedCO} & SVM-RBF  & $75.60 \%$ & n/a & n/a & n/a\\
\hline Zhao et al. \cite{Zhao2018AHS} & GBDT  & $86.94 \%$ & n/a & n/a & n/a \\
\hline HCT (Our Work) & Hybrid ConvNet-Transformer & $\mathbf{88.67\%}$ & $89.80 \%$ & $89.58 \%$ & $89.67 \%$ \\
\hline
\end{tabular}
  \label{tab:tab99}%
\end{table*}%

In Table \ref{tab:tab54}, the performance of our Multi-class ConvNet-Transformer model dedicated to PD Staging is compared to the work of Veeraragavan et al. \cite{Veeraragavan2020ParkinsonsDD}, which is, to the best of our knowledge, the only work classifying the three levels of PD severity (H\&Y=2, H\&Y=2.5, and H\&Y=3) and treating them separately from the healthy class for severity prediction as we do. Indeed, in this experiment, only PD patients are considered. The results were obtained through a 10-fold cross-validation method.
We can see that our Multi-Class ConvNet-Transformer algorithm outperforms the Feed Forward Network (FFN) algorithm used in \cite{Veeraragavan2020ParkinsonsDD} in terms of accuracy, recall, and F1-score. However, the FFN algorithm has a slightly higher precision score. Overall, our Multi-Class ConvNet-Transformer model is shown to be a little more effective for assessing the severity of Parkinson's disease. Note that our Multi-class ConvNet-Transformer model used to detect the stage of PD for parkinsonian patients was trained on the 210 parkinsonian walks divided into the three stages of the disease. The confusion matrix obtained in Figure \ref{FIG:102} shows that the Multi-class ConvNet-Transformer model achieved high accuracy for all three classes. However, the matrix also reveals some confusion between the different classes, with some samples from the H\&Y 2 class being classified as H\&Y 2.5 and vice versa. Nonetheless, the overall results demonstrate the potential of our Multi-class ConvNet-Transformer model in accurately assessing PD severity.

\subsubsection{Overall two-step method results}

\begin{figure}[t]
     \centering
		\includegraphics[scale=0.45]{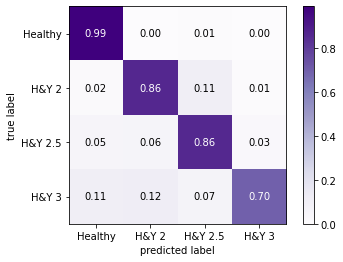}
	\caption{Confusion matrix of the overall algorithm}
	\label{FIG:97}
\end{figure}

In Table \ref{tab:tab99}, we compare the results of our overall two-step framework with those of other methods used to diagnose PD considering the four classes (Healthy, H\&Y=2, H\&Y=2.5, and H\&Y=3). These methods include Support Vector Machine with Radial Basis Function kernel (SVM-RBF)\cite{Caramia2018IMUBasedCO} and Gradient Boosting Decision Tree (GBDT)\cite{Zhao2018AHS}. The H\&Y Scale was employed in the assessment. The results show that our Hybrid ConvNet-Transformer algorithm obtained the best overall accuracy of 88.67\%. This suggests that the ConvNet-Transformer approach improves PD diagnosing by exploiting ConvNet and Transformer architectures while dividing the diagnosis process into two sub-problems of detection followed by staging. ConvNet architectures are efficient in capturing and learning local information, whereas Transformer architectures are particularly suitable for processing sequential and spatial data. Our ConvNet-Transformer model effectively collected both spatial and local information due to these two types of architectures. Figure \ref{FIG:97} shows the detailed confusion matrix resulting from our two-step classification approach. The relatively low accuracy for class 3 could be explained by dataset imbalance. In fact, the dataset contains more data for H\&Y stages 2 and 2.5 than for H\&Y stage 3. It should be noted that this observation is consistent across different SOTA methods \cite{Veeraragavan2020ParkinsonsDD, Nguyen2022TransformersF1}. 

\section{Conclusion}
\label{conclusion}
We proposed a novel two-step framework for the detection and severity assessment of Parkinson's disease (PD) using VGRF data. The proposed method achieved an accuracy of 97\% for PD detection and 87\% for severity assessment on the H\&Y scale, which improves the results compared to previous studies. Our Hybrid ConvNet-Transformer architecture takes advantage of both ConvNet and Transformer models, where the ConvNet part of the architecture is able to extract local features from the data and the Transformer part allows to capture temporal and spatial dependencies. The growing availability of biomedical sensors presents significant opportunities to widely implement our method, particularly for monitoring gait abnormalities in elderly populations.
\section*{Acknowledgment}
We acknowledge the support of the Natural Sciences and Engineering Research Council of Canada (NSERC), [Funding references: RGPIN-2020-04633 and RGPIN-2020-04937].

\bibliographystyle{IEEEtran}
\bibliography{bibliography}

\begin{thebibliography}{10}
\providecommand{\url}[1]{#1}
\csname url@samestyle\endcsname
\providecommand{\newblock}{\relax}
\providecommand{\bibinfo}[2]{#2}
\providecommand{\BIBentrySTDinterwordspacing}{\spaceskip=0pt\relax}
\providecommand{\BIBentryALTinterwordstretchfactor}{4}
\providecommand{\BIBentryALTinterwordspacing}{\spaceskip=\fontdimen2\font plus
\BIBentryALTinterwordstretchfactor\fontdimen3\font minus \fontdimen4\font\relax}
\providecommand{\BIBforeignlanguage}[2]{{%
\expandafter\ifx\csname l@#1\endcsname\relax
\typeout{** WARNING: IEEEtran.bst: No hyphenation pattern has been}%
\typeout{** loaded for the language `#1'. Using the pattern for}%
\typeout{** the default language instead.}%
\else
\language=\csname l@#1\endcsname
\fi
#2}}
\providecommand{\BIBdecl}{\relax}
\BIBdecl

\bibitem{Hughes1994ParkinsonsDA}
R.~C. Hughes, ``Parkinson's disease and its management,'' \emph{BMJ}, vol. 308, p. 281, 1994.

\bibitem{Hoehn1998ParkinsonismOP}
M.~M. Hoehn and M.~D. Yahr, ``Parkinsonism: onset, progression, and mortality. 1967.'' \emph{Neurology}, vol. 57 10 Suppl 3, pp. S11--26, 1998.

\bibitem{Fahn1987UnifiedPD}
\BIBentryALTinterwordspacing
S.~Fahn, ``Unified parkinson's disease rating scale,'' 1987. [Online]. Available: \url{https://api.semanticscholar.org/CorpusID:141435629}
\BIBentrySTDinterwordspacing

\bibitem{Vilda2017ParkinsonDD}
P.~G. Vilda, J.~Mekyska, J.~M. Ferr{\'a}ndez, D.~Palacios-Alonso, A.~G{\'o}mez-Rodellar, M.~V.~R. Biarge, Z.~Galaz, Z.~Smekal, I.~Eliasova, M.~Kostalova, and I.~Rektorov{\'a}, ``Parkinson disease detection from speech articulation neuromechanics,'' \emph{Frontiers in Neuroinformatics}, vol.~11, 2017.

\bibitem{Amato2021AnAF}
F.~Amato, L.~Borz{\`i}, G.~Olmo, and J.~R. Orozco-Arroyave, ``An algorithm for parkinson’s disease speech classification based on isolated words analysis,'' \emph{Health Information Science and Systems}, vol.~9, 2021.

\bibitem{Zhao2018AHS}
A.~Zhao, L.~Qi, J.~Li, J.~Dong, and H.~Yu, ``A hybrid spatio-temporal model for detection and severity rating of parkinson's disease from gait data,'' \emph{Neurocomputing}, vol. 315, pp. 1--8, 2018.

\bibitem{Ertugrul2016DetectionOP}
{\"O}.~F. Ertugrul, Y.~Kaya, R.~Tekin, and M.~N. Almali, ``Detection of parkinson's disease by shifted one dimensional local binary patterns from gait,'' \emph{Expert Syst. Appl.}, vol.~56, pp. 156--163, 2016.

\bibitem{Wu2017MeasuringSF}
Y.~Wu, P.~Chen, X.~Luo, M.~Wu, L.~Liao, S.~Yang, and R.~M. Rangayyan, ``Measuring signal fluctuations in gait rhythm time series of patients with parkinson's disease using entropy parameters,'' \emph{Biomed. Signal Process. Control.}, vol.~31, pp. 265--271, 2017.

\bibitem{Guo2022DetectionAA}
Y.~Guo, J.~Yang, Y.~Liu, X.~Chen, and G.-Z. Yang, ``Detection and assessment of parkinson's disease based on gait analysis: A survey,'' \emph{Frontiers in Aging Neuroscience}, vol.~14, 2022.

\bibitem{Perumal2016GaitAT}
S.~V. Perumal and R.~Sankar, ``Gait and tremor assessment for patients with parkinson's disease using wearable sensors,'' \emph{ICT Express}, vol.~2, pp. 168--174, 2016.

\bibitem{Aurolu2018ParkinsonsDM}
T.~Aşuroğlu, K.~Açıcı, Çağatay Berke~Erdaş, M.~K. Toprak, H.~Erdem, and H.~Oğul, ``Parkinson's disease monitoring from gait analysis via foot-worn sensors,'' \emph{Biocybernetics and Biomedical Engineering}, vol.~38, pp. 760--772, 2018.

\bibitem{Machi2019Deep1F}
I.~E. Ma{\^a}chi, G.-A. Bilodeau, and W.~Bouachir, ``Deep 1d-convnet for accurate parkinson disease detection and severity prediction from gait,'' \emph{Expert Syst. Appl.}, vol. 143, 2019.

\bibitem{Nguyen2022TransformersF1}
D.~M. Nguyen, M.~Miah, G.-A. Bilodeau, and W.~Bouachir, ``Transformers for 1d signals in parkinson’s disease detection from gait,'' \emph{2022 26th International Conference on Pattern Recognition (ICPR)}, pp. 5089--5095, 2022.

\bibitem{Veeraragavan2020ParkinsonsDD}
S.~Veeraragavan, A.~A. Gopalai, D.~Gouwanda, and S.~A. Ahmad, ``Parkinson’s disease diagnosis and severity assessment using ground reaction forces and neural networks,'' \emph{Frontiers in Physiology}, vol.~11, 2020.

\bibitem{Vaswani2017AttentionIA}
A.~Vaswani, N.~M. Shazeer, N.~Parmar, J.~Uszkoreit, L.~Jones, A.~N. Gomez, L.~Kaiser, and I.~Polosukhin, ``Attention is all you need,'' \emph{ArXiv}, vol. abs/1706.03762, 2017.

\bibitem{LeCun1999ObjectRW}
Y.~LeCun, P.~Haffner, L.~Bottou, and Y.~Bengio, ``Object recognition with gradient-based learning,'' in \emph{Shape, Contour and Grouping in Computer Vision}, 1999.

\bibitem{Physionet}
``Physionet dataset,'' \url{https://www.physionet.org/content/gaitpdb/1.0.0/.}

\bibitem{Yogev2005DualTG}
G.~Yogev, N.~Giladi, C.~Peretz, S.~Springer, E.~S. Simon, and J.~M. Hausdorff, ``Dual tasking, gait rhythmicity, and parkinson's disease: Which aspects of gait are attention demanding?'' \emph{European Journal of Neuroscience}, vol.~22, 2005.

\bibitem{Hausdorff2007RhythmicAS}
J.~M. Hausdorff, J.~Lowenthal, T.~Herman, L.~Gruendlinger, C.~Peretz, and N.~Giladi, ``Rhythmic auditory stimulation modulates gait variability in parkinson's disease,'' \emph{European Journal of Neuroscience}, vol.~26, 2007.

\bibitem{FrenkelToledo2005TreadmillWA}
S.~Frenkel-Toledo, N.~Giladi, C.~Peretz, T.~Herman, L.~Gruendlinger, and J.~M. Hausdorff, ``Treadmill walking as an external pacemaker to improve gait rhythm and stability in parkinson's disease,'' \emph{Movement Disorders}, vol.~20, 2005.

\bibitem{Pratama2020TrainableAF}
K.~Pratama and D.-K. Kang, ``Trainable activation function with differentiable negative side and adaptable rectified point,'' \emph{Applied Intelligence}, vol.~51, pp. 1784--1801, 2020.

\bibitem{Caramia2018IMUBasedCO}
C.~Caramia, D.~Torricelli, M.~Schmid, A.~Mu{\~n}oz-Gonzalez, J.~Gonz{\'a}lez-Vargas, F.~Grandas, and J.~L. Pons, ``Imu-based classification of parkinson's disease from gait: A sensitivity analysis on sensor location and feature selection,'' \emph{IEEE Journal of Biomedical and Health Informatics}, vol.~22, pp. 1765--1774, 2018.

\end{thebibliography}

\end{document}